\documentclass[conference]{IEEEtran}
\IEEEoverridecommandlockouts
\usepackage{cite}
\usepackage{pifont}
\usepackage{amsmath,amssymb,amsfonts}
\usepackage{algorithmic}
\usepackage{graphicx}
\usepackage{textcomp}
\usepackage[table,dvipsnames]{xcolor}
\usepackage{multicol}
\usepackage{multirow}
\usepackage{url}
\usepackage{hyperref}
\usepackage{booktabs}
\newcommand{\grayrow}{\rowcolor[gray]{0.9}}
\definecolor{darkgreen}{RGB}{60,180,75}
\def\BibTeX{{\rm B\kern-.05em{\sc i\kern-.025em b}\kern-.08em
    T\kern-.1667em\lower.7ex\hbox{E}\kern-.125emX}}
\begin{document}

\title{Non-Contact Health Monitoring During Daily Personal Care Routines\\
}

\author{
Xulin Ma$^{1*}$, Jiankai Tang$^{2,3*}$, Zhang Jiang$^{2}$, Songqin Cheng$^{2}$, Yuanchun Shi$^{1,2}$,\\ Dong Li$^{1}$, Xin Liu$^{4}$, Daniel McDuff$^{4}$, Xiaojing Liu$^{1\dagger}$, Yuntao Wang$^{1,2,3\dagger}$\\
\small $^{*}$Co-first Author, $^{\dagger}$Co-corresponding Author \\
1 Department of Computer Technology and Applications, Qinghai University, Qinghai, 810016. liuxj@qhu.edu.cn\\
2 National Key Laboratory of Human Factors Engineering, Beijing, 100094. yuntaowang@tsinghua.edu.cn\\
3 Department of Computer Science and Technology, Tsinghua University, Beijing, 100084. tjk24@mails.tsinghua.edu.cn\\
4 Paul G. Allen School of Computer Science \& Engineering, University of Washington, Seattle WA, 98195, xliu0@cs.washington.edu%
}

\maketitle

\begin{abstract}
    
Remote photoplethysmography (rPPG) enables non-contact, continuous monitoring of physiological signals and offers a practical alternative to traditional health sensing methods. Although rPPG is promising for daily health monitoring, its application in long-term personal care scenarios—such as mirror-facing routines in high-altitude environments—remains challenging due to ambient lighting variations, frequent occlusions from hand movements, and dynamic facial postures. To address these challenges, we present the \textbf{L}ong-term \textbf{A}ltitude \textbf{D}aily \textbf{H}ealth (\textbf{LADH}) dataset, the first long-term rPPG dataset containing 240 synchronized RGB and infrared (IR) facial videos from 21 participants across five common personal care scenarios, along with ground-truth PPG, respiration, and blood oxygen signals. Our experiments demonstrate that combining RGB and IR video inputs improves the accuracy and robustness of non-contact physiological monitoring, achieving a mean absolute error (MAE) of 4.99 BPM in heart rate estimation. Furthermore, we find that multi-task learning enhances performance across multiple physiological indicators simultaneously. Dataset and code are open at 
\url{https://github.com/McJackTang/FusionVitals}.
\end{abstract}


\section{Introduction}

Traditional methods for measuring vital signs, such as heart rate (HR), blood oxygen saturation (SpO$_2$), and respiratory rate (RR), often require bulky or intrusive equipment, which limits their practicality for continuous monitoring in daily life~\cite{tang2023alpha,tang2025dataset}. Remote photoplethysmography (rPPG) provides a non-invasive alternative by using cameras to detect subtle changes in skin reflectance, enabling less obtrusive monitoring in everyday settings~\cite{mcduff2023camera}. Despite these advantages, rPPG faces challenges in real-world environments, where physiological signals can be affected by factors such as clothing, hair, cosmetics, head movement, and changes in ambient lighting. This study focuses on addressing these challenges in daily personal care scenarios.

Although public benchmark datasets have contributed significantly to the advancement of rPPG research, most do not address the specific challenges found in daily personal care scenarios~\cite{liu2024spiking}. For example, commonly used rPPG datasets such as PURE~\cite{stricker2014non}, UBFC-rPPG~\cite{ubfcrppg}, and SUMS~\cite{liu2024summit} are mainly collected under resting-state conditions and may not capture the physiological variations that occur during routine personal care activities.

\begin{figure}[htbp]
\centerline{\includegraphics[width=0.8\linewidth]{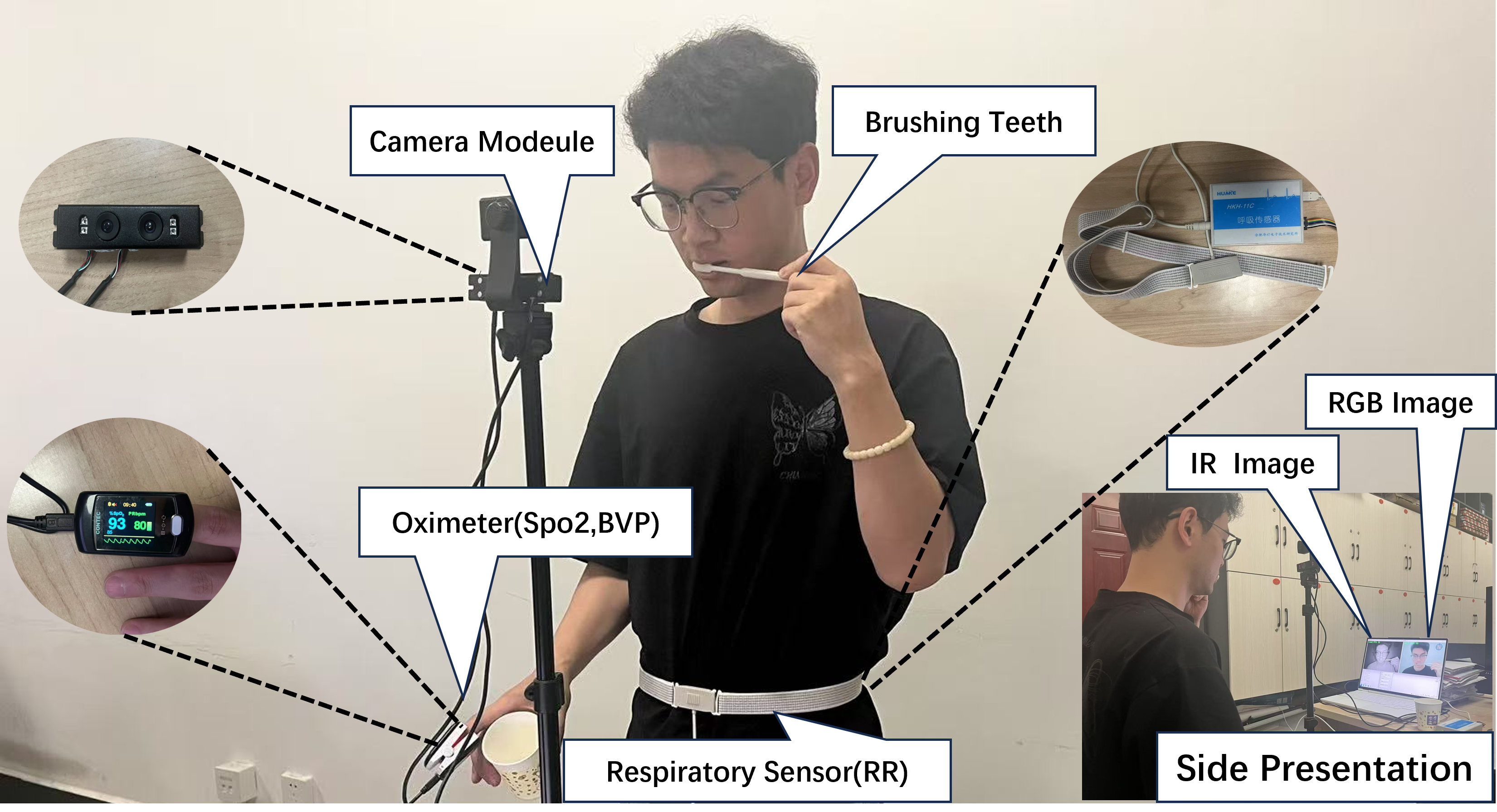}}
\vspace{-0.8em} 
\caption{The experimental setup of data collection while participants are brushing teeth.}
\vspace{-1.5em}   
\label{fig:Collection}
\end{figure}

To address these challenges, we present \textbf{L}ong-term \textbf{A}ltitude \textbf{D}aily \textbf{H}ealth (\textbf{LADH}) dataset, a multi-modal biosensing dataset for daily personal care activities. LADH contains 240 synchronized RGB and IR facial videos collected across five scenarios: sitting at rest, sitting during personal care (toothbrushing/hair-combing with facial occlusions), standing at rest, standing during personal care, and post-exercise. Eleven participants contributed data over a continuous \textbf{10-day} period, with all recordings synchronized at millisecond precision with ground-truth physiological signals from pulse oximeters and respiratory bands~\cite{liu2024summit}.

Our evaluation using both subject-independent and day-wise partitioning reveals that day-wise partitioning enhances measurement accuracy, while RGB and IR video fusion with multi-task learning significantly improves physiological signal prediction despite frequent facial occlusions.

Our main contributions include:
\begin{itemize}
\item The first long-term rPPG dataset featuring synchronized RGB and IR videos with ground-truth physiological measurements across daily personal care scenarios with realistic facial occlusions.
\item Demonstration that multi-modal fusion of RGB and IR inputs improves non-contact physiological monitoring accuracy (achieving 4.99 BPM MAE) despite facial occlusions during toothbrushing and hair-combing.
\item Evidence that multi-task learning enhances performance across multiple physiological indicators simultaneously in both heart and respiratory rate estimation.
\end{itemize}

\begin{table}[t]
\caption{Dataset Comparison}
\vspace{-0.8em} 
\centering
\setlength{\tabcolsep}{2pt}
\resizebox{\columnwidth}{!}{%
\renewcommand{\arraystretch}{1.2}
\begin{tabular}{cccccc}
\hline
Dataset & Videos & Camera-Position & Vitals  & Long-term & Obscured   \\
\hline
\hline
PURE~\cite{stricker2014non} & 40 & Face & PPG/SpO$_2$  & \ding{55} & \ding{55}  \\
UBFC-rPPG~\cite{ubfcrppg} & 42 & Face & PPG  & \ding{55} & \ding{55}  \\
MMPD~\cite{tang2023mmpd} & 660 & Face & PPG  & \ding{55} & \ding{55}   \\
SUMS~\cite{liu2024summit} & 80 & Face+Finger & PPG/SpO$_2$/RR  & \ding{55}  & \ding{55} \\
\hline
LADH & 240 & Face(RGB+IR) & PPG/SpO$_2$/RR &   \ding{51} & \ding{51} \\
\hline
\end{tabular}%
}
\label{tab: dataset}
\vspace{-1.5em} 
\end{table}

\section{Related Works}
\subsection{Physiological Sensing in Daily Personal Care Scenarios}

With the development of non-contact physiological sensing technologies, daily personal care activities such as face washing, mirror viewing, and tooth brushing have become practical contexts for health monitoring due to their routine and accessible nature~\cite{tang2024camera}. Regular monitoring of vital signs, including heart rate, blood pressure, and blood oxygen saturation, may support early detection and prevention of various health conditions~\cite{talukdar2022evaluation}. Unlike controlled laboratory settings, these everyday activities provide opportunities for unobtrusive physiological data collection in real-world environments.

Previous studies have demonstrated the feasibility of smart mirror systems for non-contact, real-time heart rate measurement without external sensors~\cite{poh2011medical,dobrovoljski2017smartmirror}. These results indicate that non-contact health monitoring systems can offer a passive, user-friendly, and privacy-preserving approach in daily personal care scenarios. However, there is limited research on the robustness of such systems in long-term, comprehensive user studies. The LADH dataset aims to address this gap by focusing on daily personal care routines in the real world.

\subsection{Physiological Sensing Datasets}
Many widely used remote photoplethysmography (rPPG) datasets—such as PURE~\cite{stricker2014non}, UBFC-rPPG~\cite{ubfcrppg}, MMPD~\cite{tang2023mmpd}, and SUMS~\cite{liu2024summit}—have been collected under controlled laboratory conditions, typically involving static seated positions or limited facial movement. As a result, these datasets may not fully capture the dynamic variations present in daily personal care activities, such as tooth brushing or mirror-facing routines.

To address this limitation, we developed the LADH dataset to support the evaluation of rPPG methods in everyday personal care scenarios. LADH contains synchronized RGB and IR facial videos, along with ground-truth physiological signals, collected from 21 participants over 10 days in real-world environments. Compared to existing datasets, LADH introduces challenges such as facial occlusion and varying indoor lighting, providing a resource for studying the robustness of non-contact physiological signal measurement. Table~\ref{tab: dataset} summarizes several related datasets for comparison.

\begin{figure}[htbp]
\centering
\includegraphics[width=0.9\linewidth]{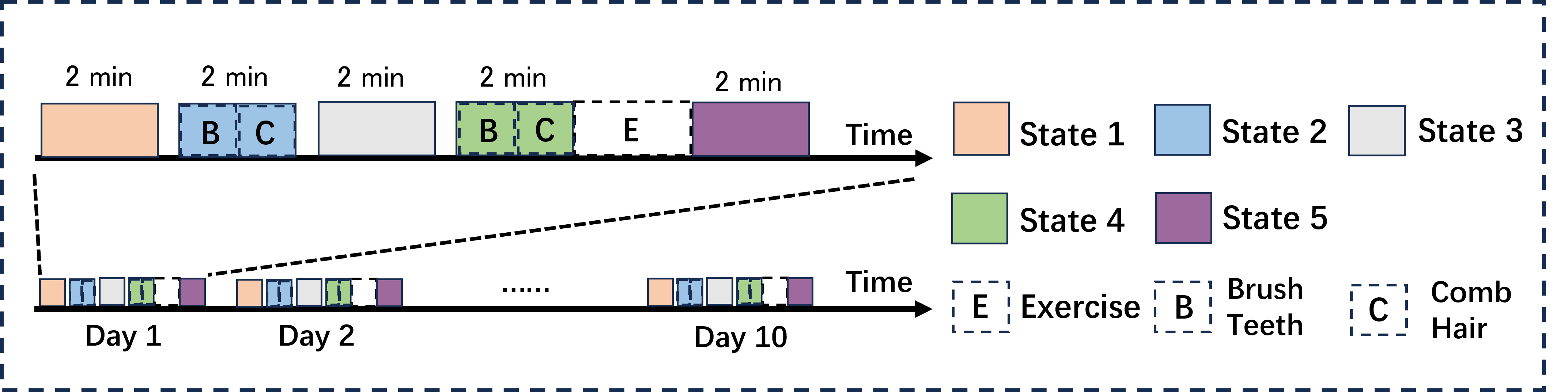}
\vspace{-0.5em} 
\caption{A visual illustration of our daily data collection protocol. Participants
have different activities across states.}
\vspace{-1.5em}   
\label{fig:states}
\end{figure}

\section{Dataset}
Approved by the Institutional Review Board, we recruited 21 graduate students (aged 23–28) as participants and conducted data collection in a simulated daily personal care scenario. The WN-12207K3321SM290 camera module was used to capture participants’ facial videos in both RGB and IR modalities. As shown in Figure~\ref{fig:Collection}, physiological ground-truth signals were recorded using a CMS50E pulse oximeter for photoplethysmography (PPG) and SpO$_2$, and an HKH-11C respiratory sensor to monitor breathing patterns. Video recordings were acquired at a resolution of 640×480 pixels and a frame rate of 30 frames per second (FPS). PPG signals were sampled at 20 Hz, while RR signals were recorded at 50 Hz. All video frames and physiological signals were synchronously collected at millisecond-level precision using a data acquisition platform based on the PhysRecorder~\cite{wang2024camera}. 

\subsection{Data Collection}
The data collection protocol is illustrated in Figure \ref{fig:states}. Data were collected from 21 participants in five scenarios, where 11 subjects participated in a 10-day longitudinal study, while the remaining 10 contributed single-day recordings. For the seated resting condition (state 1), participants wore an HKH-11C respiratory sensor on their abdomen and a CMS50E pulse oximeter on their left index finger while sitting upright facing the camera. Participants maintained minimal movement with a fixed gaze at the camera during the two-minute recording period.

In state 2, participants remained seated with the same equipment while performing toothbrushing and hair-combing activities for two minutes. State 3 involved a standing resting position with the same sensor configuration, also recorded for two minutes. In state 4, participants repeated the toothbrushing and hair-combing actions while standing for two minutes. After completing these four conditions, participants engaged in moderate physical exercise (squats, high knees, or breath-holding) to induce physiological changes. Immediately following exercise, participants were recorded for an additional two minutes in a seated position (state 5). The exercise component was included to produce variations in SpO$_2$ levels. 

\begin{figure}[tp]
\centerline{\includegraphics[width=1\linewidth]{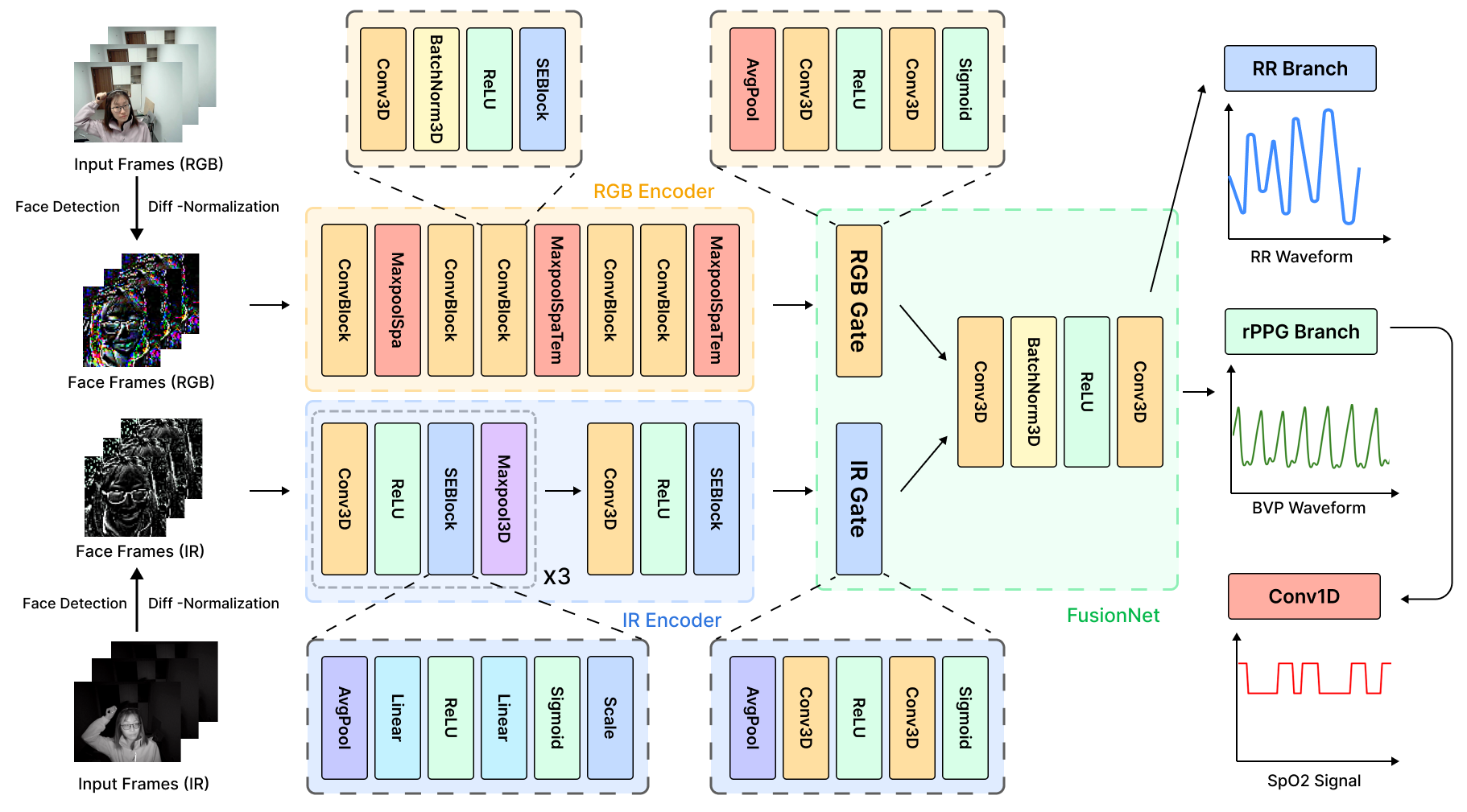}}

\vspace{-0.5em}   
\caption{FusionPhys Model with Input frames of facial RGB and facial IR. PPG, RR and SpO2 estimation tasks are trained simultaneously with a combined loss.}
\vspace{-1.5em}   
\label{fig:model}
\end{figure}

\section{Method}

\subsection{Input embedding}
Building upon PhysNet~\cite{physnet,liu2024summit}, our FusionPhysNet introduces an input embedding strategy that processes facial video from two modalities: RGB and IR. As shown in Figure~\ref{fig:model}, both video streams are pre-processed into frame sequences and passed through a shared 3D convolutional encoder for spatiotemporal feature extraction. By integrating features from both modalities, the model achieves a more comprehensive understanding of an individual’s physiological state, effectively combining superficial and deeper physiological signals. This multimodal fusion framework significantly enhances the model’s performance in complex real-world scenarios and establishes a more accurate and reliable foundation for vital sign estimation, including heart rate and blood oxygen saturation.

\subsection{Neural Network Model}
The overall architecture of the FusionPhysNet is illustrated in Figure \ref{fig:model}. We extend the PhysNet~\cite{physnet} backbone with a modality-aware fusion mechanism in FusionNet. Specifically, a gated feature selection strategy adaptively modulates the contribution of each modality based on global contextual representations.  This design enables the model to dynamically emphasize the more informative modality under varying environmental conditions (e.g., changes in illumination), enhancing the robustness and generalizability of the physiological signal estimation framework.

\begin{table}[bp]
\caption{Intra-Dataset and Inter-Dataset Experiment}
\vspace{-0.8em} 
\centering
\resizebox{\columnwidth}{!}{%
\renewcommand{\arraystretch}{1.2} 
\begin{tabular}{c|cccccc}
\hline
\toprule {Training Set}&  \multicolumn{2}{c}{LADH} &  \multicolumn{2}{c}{SUMS} &  \multicolumn{2}{c}{PURE}\\
\textbf{Test Set}& MAE$\downarrow$ & MAPE$\downarrow$ &MAE$\downarrow$ & MAPE$\downarrow$ & MAE$\downarrow$ & MAPE$\downarrow$ \\ 
\midrule
\midrule
LADH & 8.15 & 9.19 & 16.93 & 18.20 & 17.00 & 18.78 \\
SUMS & 11.23 & 15.45 & 3.36 & 3.84 & 14.95 & 17.11 \\
PURE & 8.10 & 8.83 & 7.97 & 8.87 & 0.59 & 0.77 \\
\bottomrule[1.5pt]
\end{tabular}%
}
\footnotesize
      The table shows the cross-dataset experimental results of the LADH, SUMS~\cite{liu2024summit}, and PURE~\cite{stricker2014non} datasets on the \textbf{PhysNet~\cite{physnet}} model.
\label{tab:Comparative_Experiment}
\end{table}

\subsection{Joint-Training}
Inspired by the multitask training framework~\cite{liu2024summit}, we incorporate RR estimation into the joint training mechanism, allowing simultaneous optimization of HR, SpO$_2$ and RR prediction tasks. Furthermore, we introduce a novel loss function specifically designed to jointly handle the objectives of HR, SpO$_2$ and RR estimation. By optimizing these three targets concurrently during training, the model is guided to learn a more comprehensive and informative feature representation, thus improving its capacity to capture complex physiological dynamics and improving the robustness of vital sign prediction. It is formulated as:
\begin{align*}
\text{Loss} &= \mathrm{MSE}_{\mathrm{BVP}} + \mathrm{MSE}_{\mathrm{RR}} \\
&\quad + 0.002 \times \mathrm{MSE}_{\mathrm{SpO_2}} \times \left(100 - \mathrm{Mean}(\mathrm{SpO_2})\right)
\end{align*}
where \( \text{MSE}_{\text{BVP}} \) represents the mean squared error of the blood volume pulse (BVP) signal, \( \text{MSE}_{\text{RR}} \) represents the mean squared error of the RR signal, and \( \text{MSE}_{\text{SpO$_2$}} \) represents the mean squared error of the SpO$_2$ signal. This formulation aims to balance the precision of the BVP, RR and SpO2 measurements, leading to a more robust model for multimodal physiological signal measurement.

\section{Results and Findings}

All experiments were conducted using the improved training framework of rPPG-Toolbox~\cite{liu2024rppg}. The environment configuration included PyTorch 2.2.2+cuda12.1, Python 3.8, and NVIDIA A100 GPU. The experiments were performed with a learning rate of 9e-3, 30 epochs, and a batch size of 16.

We validated the effectiveness of LADH through both intra- and inter-dataset experiments using PhysNet~\cite{physnet}. As shown in Table~\ref{tab:Comparative_Experiment}, LADH achieves reasonable intra-dataset accuracy (MAE=8.15 BPM) and maintains competitive cross-dataset performance when tested on SUMS and PURE, outperforming SUMS in cross-transfer to PURE (MAE 8.10 vs. 14.95). These results highlight LADH’s dual role: it provides a strong baseline within its own domain while also exposing the inherent challenges of generalization across different datasets, underscoring its value as a benchmark for robust rPPG evaluation.

We then conducted two sets of experiments based on different dataset partitioning strategies within LADH: subject-wise partitioning and day-wise partitioning. We established different training tasks, including individual training for HR, SpO$_2$ and RR  under multimodal input, joint training for HR-SpO$_2$-RR under multimodal input, joint training for HR-SpO$_2$-RR under RGB-only video input and IR-only video input.

\begin{table}[tp]
\caption{Results of HR-SpO$_2$-RR Multi-Task Training by Subject}
\vspace{-0.8em} 
\centering
\resizebox{\columnwidth}{!}{%
\renewcommand{\arraystretch}{1.2} 
\begin{tabular}{c|cc|cc|cc}
\hline
\toprule \multirow{2}{*}{\textbf{Modality}} &  \multicolumn{2}{c}{\textbf{HR Task}} &  \multicolumn{2}{c}{\textbf{SpO2 Task}} &  \multicolumn{2}{c}{\textbf{RR Task}}\\
& MAE$\downarrow$ & MAPE$\downarrow$ &MAE$\downarrow$ & MAPE$\downarrow$ & MAE$\downarrow$ & MAPE$\downarrow$ \\ 
\midrule
\midrule
Both(Single Task)& 9.02&10.99 & \underline{1.10}&\underline{1.19} & 2.25&10.16 \\
RGB(Multi Task) & 9.34&12.08 & 1.29&1.39 & 3.08&13.78 \\
IR(Multi Task) & 12.99&15.73 & 1.23&1.33 & 2.41&11.20 \\
\grayrow
Both(Multi Task) & \textbf{\underline{7.12}}&\textbf{\underline{8.93}} & \textbf{1.14}&\textbf{1.23} & \textbf{\underline{1.43}}&\textbf{\underline{6.53}} \\
\midrule
\textsc{Gains} & \textcolor{darkgreen}{\texttt{+}\textbf{1.90}} & \textcolor{darkgreen}{\texttt{+}\textbf{2.06}}  & \textcolor{SkyBlue}{\texttt{-}\textbf{0.04}} & \textcolor{SkyBlue}{\texttt{-}\textbf{0.04}}  & \textcolor{darkgreen}{\texttt{+}\textbf{0.82}} & \textcolor{darkgreen}{\texttt{+}\textbf{3.63}}  \\
\bottomrule[1.5pt]
\end{tabular}%
}
\\
\footnotesize
       MAE = Mean Absolute Error in HR estimation (Beats/Min), MAPE = Mean Percentage Error (\%). \underline{underline} means the best performance. \textbf{Gains} denote the improvements brought by multimodal and multitask learning.
\label{tab:results-subject}
\end{table}

\textbf{In the subject-wise partitioning experiment, multimodal fusion with joint training outperforms single-modality and single-task approaches, particularly for HR and RR estimation.}
The dataset was partitioned such that data from 8 subjects were used for training, 3 subjects for validation, and an additional dataset from 10 individuals was reserved for testing. The results indicated significant improvements in the MAE for HR, which decreased from 9.02 to 7.12, reflecting a 21.06\% error reduction, and for RR, which decreased from 2.25 to 1.43, reflecting a 36.44\% error reduction. This suggests that multimodal fusion and joint training are more effective for periodic tasks like HR and RR, while SpO$_2$ does not exhibit clear periodic fluctuations and is inferred through indirect signals.

\textbf{In the day-wise partitioning experiment, multimodal fusion with joint training improves HR estimation, and multitask learning benefits SpO$_2$ and RR estimation.}
In this experiment, data collected over 10 days were split into 7 days for training, 2 days for validation, and 1 day for testing. The results showed that for HR estimation, multimodal fusion with joint training outperformed single-modality and single-task approaches, reducing MAE from 5.23 to 4.99 (a 4.59\% error reduction). In IR-based joint training, errors for SpO$_2$ and RR were reduced by 2.29\% and 41.25\%, respectively. This highlights the effectiveness of multimodal fusion for HR and multitask learning for SpO$_2$ and RR.

\textbf{Comparison of the subject-wise and day-wise experiments illustrates how day-wise analysis can improve the adaptability of models to individual user data.}
While the subject-wise experiment shows strong performance for periodic tasks through multimodal fusion and joint training, the day-wise experiment emphasizes the ability of the model to adapt more closely to individual data. This could indicate that, in future personalized health monitoring systems, such as a health mirror, models can better accommodate daily variations and offer more tailored results to users, enhancing the accuracy of HR, RR, and SpO$_2$ estimation on an individual level.

\begin{table}[tp]
\caption{Results of HR-SpO$_2$-RR Multi-Task Training by Day}
\vspace{-0.8em} 
\centering
\resizebox{\columnwidth}{!}{%
\renewcommand{\arraystretch}{1.2} 
\begin{tabular}{c|cc|cc|cc}
\hline
\toprule \multirow{2}{*}{\textbf{Modality}} &  \multicolumn{2}{c}{\textbf{HR Task}} &  \multicolumn{2}{c}{\textbf{SpO2 Task}} &  \multicolumn{2}{c}{\textbf{RR Task}}\\
& MAE$\downarrow$ & MAPE$\downarrow$ &MAE$\downarrow$ & MAPE$\downarrow$ & MAE$\downarrow$ & MAPE$\downarrow$ \\ 
\midrule
\midrule
Both(Single Task)& 5.23 & 5.44 & 1.31 & 1.38 & 2.57 & 13.45 \\
RGB(Multi Task) & 5.73 & 5.77 & 1.35 & 1.43 & 1.99 & 9.12 \\
IR(Multi Task) & 8.35 & 8.98 & \underline{1.28} & \underline{1.36} & \underline{1.51} & \underline{6.74} \\
\grayrow
Both(Multi Task) & \textbf{\underline{4.99}} & \textbf{\underline{5.21}} & \textbf{1.29} & \textbf{1.37} & \textbf{2.24} & \textbf{11.38} \\
\bottomrule[1.5pt]
\end{tabular}%
}
\footnotesize
      MAE = Mean Absolute Error in HR estimation (Beats/Min), MAPE = Mean Percentage Error (\%). \underline{underline} means the best performance. 
\label{tab:results-day}
\end{table}

\section{Limitations}
While this study provides valuable insights into the feasibility of non-contact health monitoring in daily personal care scenarios, it has certain limitations. The relatively small sample size and age range may not fully capture the diversity of physiological characteristics and behavioral patterns across the general population. Future research should aim to include a larger and more diverse participant pool and incorporate a broader range of real-world personal care behaviors to enhance the generalizability and robustness of the findings.

\section{Conclusion}
This study introduces the LADH dataset, the first long-term rPPG dataset with synchronized RGB, IR, and ground-truth physiological signals across realistic daily care scenarios. By releasing 240 multimodal videos with PPG, respiration, and SpO$_2$, LADH provides a valuable benchmark for robust non-contact health monitoring. We further propose FusionPhys, a modality-fusion and multi-task learning framework that improves accuracy and generalization, reducing HR and RR errors across both cross-subject and cross-day settings. FusionPhys reveals the complementary value of RGB and IR signals and the effectiveness of joint HR–SpO$_2$–RR prediction. This work lays a foundation for accurate, robust, and non-invasive health monitoring in everyday scenarios.

\section*{Acknowledgment}
This work is supported by the foundation of National Key Laboratory of Human Factors Engineering No. HFNKL2024W06, the National Natural Science Foundation of China No. 62366043 \& 62472244, Beijing Natural Science Foundation No.QY24248, the National Key R\&D Program of China No. 2024YFB4505500, No. 2024YFB4505503, the Tsinghua University Initiative Scientific Research Program No. 20257020004, and Qinghai University Research Ability Enhancement Project No. 2025KTSA05.

{
    \small
    \bibliographystyle{unsrt}
\bibliography{main}
}

\end{document}